\documentclass[10pt,twocolumn,letterpaper]{article}
\usepackage{cvpr}              

\usepackage[dvipsnames]{xcolor}

\usepackage{microtype}
\usepackage{graphicx}
\usepackage{booktabs} 
\usepackage{amssymb}
\usepackage{amsthm}
\usepackage{amsmath}
\usepackage{array} 
\usepackage{mathtools} 
\usepackage{float}
\usepackage{soul}
\usepackage{multirow}
\usepackage{IEEEtrantools}
\usepackage{bbm}
\usepackage{graphicx}
\usepackage[export]{adjustbox}
\usepackage{tabulary}
\usepackage{tabularx}
\usepackage{natbib}

\usepackage[dvipsnames]{xcolor}

\makeatletter
\@namedef{ver@everyshi.sty}{}
\makeatother
\usepackage{tikz}

\usepackage[T1]{fontenc}
\usepackage[utf8]{inputenc}
\usepackage{mathrsfs}
\usepackage{tcolorbox}

\newenvironment{tight_enumerate}{
\begin{enumerate}[leftmargin=15pt]
  \setlength{\topsep}{0pt}
  \setlength{\itemsep}{0pt}
  \setlength{\parskip}{0pt}
  \setlength{\parsep}{0pt}
}{\end{enumerate}}

\definecolor{cvprblue}{rgb}{0.21,0.49,0.74}
\usepackage[pagebackref,breaklinks,colorlinks,citecolor=cvprblue]{hyperref}

\def\paperID{24} 
\def\confName{CVPR}
\def\confYear{2024}

\title{Robust Disaster Assessment from Aerial Imagery Using \\ Text-to-Image Synthetic Data}

\author{
     \textbf{Tarun Kalluri}$^{1}$\thanks{Work done during TK's internship at Google.} \quad
     \textbf{Jihyeon Lee}$^{2}$ \quad
     \textbf{Kihyuk Sohn}$^{2}$ \quad
     \textbf{Sahil Singla}$^{2}$ \quad \\
     \textbf{Manmohan Chandraker}$^{1}$ \quad
     \textbf{Joseph Xu}$^{2}$ \quad
     \textbf{Jeremiah Liu}$^{2}$ \\
     $^1$UC San Diego\quad
     $^2$Google Research\\
}

\begin{document}
\maketitle

\begin{abstract}
We present a simple and efficient method to leverage emerging text-to-image generative models in creating large-scale synthetic supervision for the task of damage assessment from aerial images. While significant recent advances have resulted in improved techniques for damage assessment using aerial or satellite imagery, they still suffer from poor robustness to domains where manual labeled data is unavailable, directly impacting post-disaster humanitarian assistance in such under-resourced geographies. Our contribution towards improving domain robustness in this scenario is two-fold. Firstly, we leverage the text-guided mask-based image editing capabilities of generative models and build an efficient and easily scalable pipeline to generate thousands of post-disaster images from low-resource domains. Secondly, we propose a simple two-stage training approach to train robust models while using manual supervision from different source domains along with the generated synthetic target domain data. We validate the strength of our proposed framework under cross-geography domain transfer setting from xBD and SKAI images in both single-source and multi-source settings, achieving significant improvements over a source-only baseline in each case. 
\end{abstract}

\section{Introduction}
\label{sec:introduction}

In this work, we address the issue of poor robustness caused by traditional training methods for the task of disaster assessment by generating synthetic data using guided text-to-image generation~\cite{ramesh2021zero, chang2023muse}. 
To accelerate rescue, recovery and aid routing through scalable and automated disaster assessment from images, recent methods propose efficient training paradigms using paired labeled data from before and after the disaster~\cite{gupta2019xbd, Bin2022AnEA, Nasrallah2021SciNetSM, Wu2021BuildingDD, Zhao2020BuildingDE, Weber2022Incidents1MAL}. While being instrumental in significantly improving the accuracy in damage assessment, these methods greatly rely on manual supervision for efficient performance and perform poorly when deployed in novel domains - such as new disaster types or unseen geographies. 
While unsupervised adaptation methods exist to overcome the overhead of manual annotation~\cite{benson2020assessing, bouchard2022, Li2020UnsupervisedDA}, they still require unlabeled images captured from both before and after the disaster for learning domain agnostic features. 
While readily available satellite imagery provides generalized aerial coverage for most geographic locations for pre-disaster images, the retrieval of post-disaster imagery remains a time-consuming process, hindering rapid damage assessment during critical response windows, with non-trivial domain shifts preventing cross-geographical deployment.

\begin{figure}[!tpb] 
    \centering
    \includegraphics[width=0.46\textwidth]{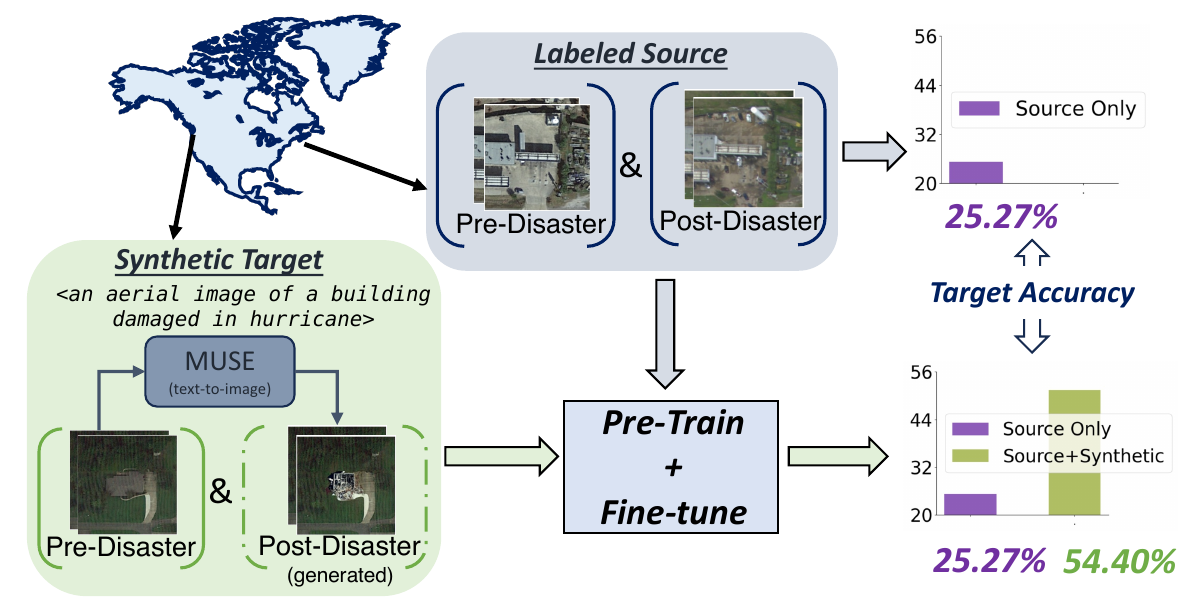} 
    \caption{{\bf Summary of our proposed pipeline.} A disaster assessment model trained using labeled data from a different domain suffers from poor accuracy due to significant distribution shifts with the target. We offer a novel way of addressing this limitation, by leveraging the recent advances in mask-based text-to-image models~\cite{chang2023muse} to generate thousands of synthetic labeled data from the target domain where only pre-disaster images are accessible. We incorporate this synthetic data along with source labeled data in a two-stage training framework to achieve significant gains on challening transfer settings from xBD~\cite{gupta2019xbd} and SKAI~\cite{lee2020assessing} datasets.} 
    \label{fig:intro_fig} 
\end{figure}

On the other hand, there has been a notable progress in the field of generating synthetic data using guided text-to-image models~\cite{sariyildiz2023fake, tian2024stablerep, zhou2023training, du2024dream}, which overcome the cumbersome manual annotation process and enable controllable data-generation at scale to train robust and data-efficient models. While a majority of these works focus on tasks like image recognition~\cite{sariyildiz2023fake, tian2024stablerep, tian2023learning}, object detection~\cite{ge2023beyond} and semantic segmentation~\cite{yang2024freemask, nguyen2024dataset}, extending these methods to suit the current setting of aerial disaster assessment is non-trivial, as it requires generating precisely synchronized pre- and post-disaster imagery of the affected area.


We present a novel framework that leverages image-guided text-conditioned generative models to synthesize large datasets of post-disaster imagery conditioned on pre-disaster imagery, which is trained efficiently using a two-stage approach by incorporating unlabeled target domain data alongside source labels.
By exploiting the mask-based image-editing abilities of transformer-based text-to-image models, we edit the pre-disaster image using a suitable text-prompt to create a corresponding synthetic post-disaster image, generating a new, synthetically labeled dataset specific to the target domain. To mitigate performance degradation caused by domain shift between generated data and real-world images, we adopt a two-stage training procedure. We first train a siamese vision-transformer~\cite{Da2022BuildingDA} using labeled data from the source domains, and subsequently fine-tune the last layer on the synthetic data from the target domain following prior work~\cite{kirichenko2022last}. 
We show the effectiveness of our framework in training robust models through experiments on several challenging transfer settings from xBD and SKAI datasets, significantly outperforming a source-only training baseline in each case.
As shown in \cref{fig:intro_fig}, while training directly using the source domain only achieves only 25\% accuracy on the target test-data, our synthetic-data augmented training achieves 54.4\%, with a non-trivial improvement of 29\% on the challenging xBD dataset. In summary, our contributions are as follows.

\begin{tight_enumerate}
    \item We offer a cost-effective way to generate training data for disaster assessment in areas lacking real-world aerial imagery, leveraging the image-editing abilities of large-scale text-to-image models.
    \item Following prior work in robustness studies~\cite{kirichenko2022last}, we devise a simple and effective two-stage training strategy to use the synthetically generated data in training along with labeled data from different source domains to achieve complementary benefits.
    \item We validate the effectiveness of our proposed framework on two benchmark datasets xBD~\cite{gupta2019xbd} and SKAI~\cite{lee2020assessing} images, obtaining significant improvements over a standard source-only baseline in both single-source (+9.8\%, +25.2\%) and multi-source (+5.33\%, +29.13\%) domain transfer settings. 
\end{tight_enumerate}

\section{Related Work}
\label{sec:related_work}

\paragraph{Disaster Assessment using Satellite Images} The task of image-based disaster assessment involves predicting the presence or extent of damage in a particular location by comparing pre and post disaster aerial or satellite imagery. 
Fueled by the availability of paired pre- and post-disaster images captured from remote-sensing satellites~\cite{gupta2019xbd, Nasrallah2021SciNetSM, lee2020assessing, Xia2022OpenEarthMapAB, rahnemoonfar2021floodnet, sirko2021continental}, several methods have been proposed to identify the damage~\cite{bai2020pyramid,weber2020building,Da2022BuildingDA,Lu2024BitemporalAT}, as well as to precisely localize the damage within the image~\cite{gupta2021rescuenet, potnis2019multi, wu2021building}. However, these approaches rely on labeled data for efficient performance, preventing their use in novel domains without incurring additional collection and annotation overheads~\cite{Xu2019BuildingDD, bouchard2022}. While domain adaptation methods exist to bridge this gap~\cite{benson2020assessing, Li2020UnsupervisedDA, Valentijn2020transferability}, they still need to access the post-disaster imagery which is difficult to acquire in a short window following a disaster. While prior works attempt generation of images using GANs~\cite{wang2021gan}, they lack the ability to generate controllable synthetic data at scale. 
Our work addresses these limitations by leveraging the advances in conditional text-to-image capabilities to generate large-scale synthetic supervision from low-resource target domains. 
We also note that while there has been significant advances in unsupervised domain adaptation for image classification~\cite{kalluri2022memsac, xu2021cdtrans, Saito_2018_CVPR, zhu2023patch} and segmentation~\cite{Lai2022DecoupleNetDN, tsai2018learning}, they are typically not applicable to expert tasks like disaster assessment through aerial imagery, preventing their direct use or comparison for our problem.

\paragraph{Creating Synthetic Data from Generative Models} Recent progress in the field of generative modeling has enabled the creation of diverse and realistic images conditioned on a variety of inputs such as text~\cite{rombach2022high, ramesh2021zero, saharia2022photorealistic, chang2023muse}, images~\cite{ruiz2023dreambooth, lu2021cigli}, layouts~\cite{zheng2023layoutdiffusion} or semantic maps~\cite{zhang2023controllable, wang2022semantic}. In particular, text-to-image synthesis enables creation of diverse visual content based on natural language prompts~\cite{kawar2023imagic, chang2023muse, ramesh2021zero, saharia2022photorealistic, yu2022scaling}. Recent works explored the use of leveraging the power of these models in generating synthetic supervision for various tasks including object recognition~\cite{sariyildiz2023fake, tian2024stablerep, zhou2023training, tian2023learning, azizi2023synthetic}, object detection~\cite{ge2023beyond, xie2023boxdiff, lin2023explore}, semantic segmentation~\cite{yang2024freemask, nguyen2024dataset}, outlier detection~\cite{du2024dream} and long-tailed robustness~\cite{yu2023diversify, shao2024diffult, shin2023fill}. 
Building upon this line of work, our work tackles image generation of post-disaster imagery in low-resource domains through localized editing of the corresponding pre-disaster images guided by suitable text prompts, showing an efficient way to improve domain robustness. 

\section{Method}

\subsection{Problem Setting} 

\begin{figure*}[!thpb] 
    \centering
    \includegraphics[width=\textwidth]{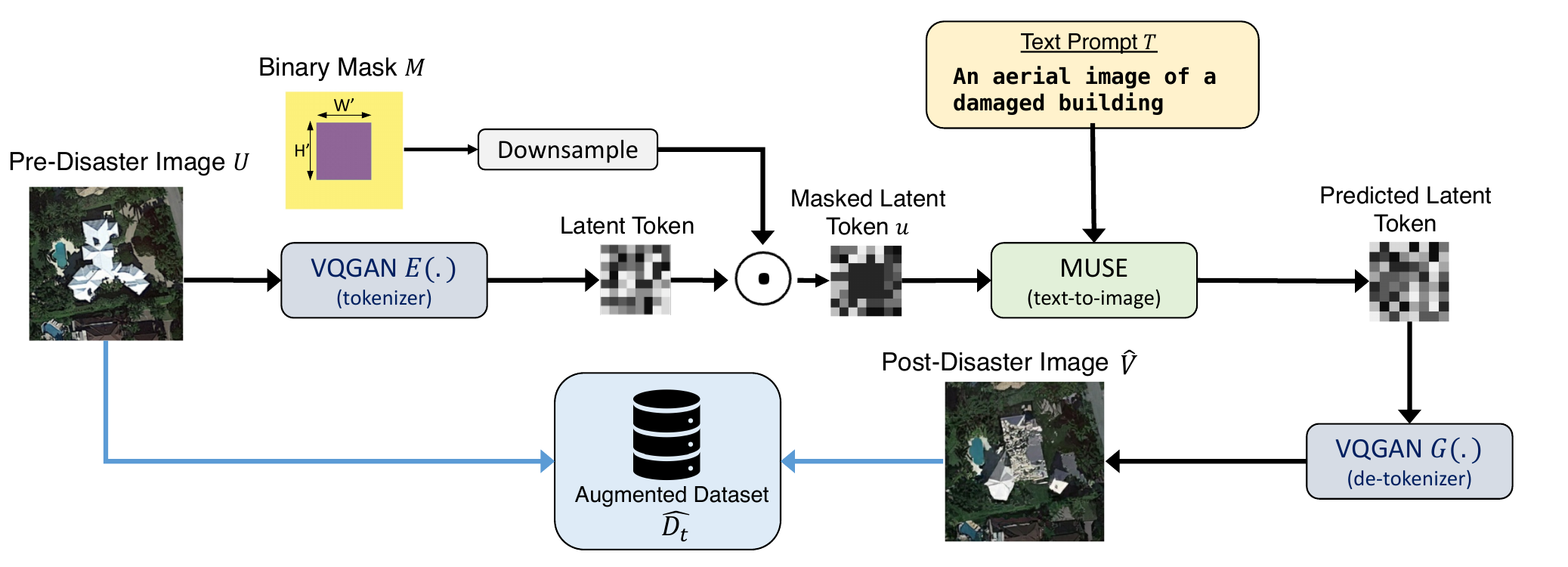} 
    \caption{{\bf Overview of the proposed synthetic data generation pipeline.} We first pass the pre-disaster image $U$ from the target domain through a pre-trained VQGAN encoder followed by a tokenizer to compute the latent token, which is then masked using a binary mask. We use the MUSE model along with a suitable text prompt $T$ to predict the output tokens from the masked tokens $u$, which are then de-tokenized to generate the post-disaster image $\hat{V}_t$. Our augmented dataset $\hat{D}_t$ now contains the input image $U_t$, generated image $\hat{V}_t$ and the binary label corresponding to the text prompt (indicating damage or no-damage).} 
    \label{fig:generation_fig} 
\end{figure*}

We now describe our problem setting of investigating domain robustness in image-based disaster assessment tasks. We denote our labeled source dataset as $\mathcal{D}_s = \{U_s^i, V_s^i, y^i\}_{i=1}^{N_s}$, where $U^i$ is the \textit{before} image (image captured before the disaster, also called pre-image), $V^i$ is the \textit{after} image (image captured after the disaster, also called the post-image) along with a binary label $y^i \in \{0,1\}$ indicating whether or not there is damage between the images due to a disaster. $N_s$ denotes the number of source images. In general, the pre and post images from the source images are paired, where the image collection is synchronized to capture the same location before and after the disasters, ensuring pixel-level correspondence between the images.
Furthermore, the target domain is denoted by $\mathcal{D}_t = \{U_t^i\}_{i=1}^{N_t}$, where $N_t$ denotes the number of (unlabeled and unpaired) target images. 
Unique to our work, we assume a \textit{zero-shot} target setting, where we only have access to the pre-images from a new domain (which could be a new geographical location or new disaster type), and neither the post images nor the damage labels are available.  
This setting is more realistic as capturing paired images immediately after a disaster and labeling them for damage incurs expensive time and manual annotation overhead in most cases, while pre-disaster images are naturally available through satellite footage.
Therefore, our main goal is to utilize labeled source images along with unlabeled pre-disaster target images to improve the performance on the target domain at test-time.

\noindent While unsupervised domain adaptation (UDA)~\cite{benson2020assessing, Li2020UnsupervisedDA} has been a standard approach to address such domain shifts for disaster assessment, UDA methods still require access to paired pre and post-disaster data during training from both source and target domains. In contrast, we only require images captured before the disaster, and automatically generate synthetic post-disaster images to facilitate robust training. 

\subsection{Background: Text-to-Image Generation} 

In our work, we adopt the MUSE~\cite{chang2023muse} model for text-to-image generation, and we provide a brief overview of that framework here. 
MUSE is a non-autoregressive model for text-conditioned synthesis capable of generating high-resolution images with fast inference speeds. In contrast to diffusion models requiring sequential decoding over several time-steps~\cite{rombach2022high, dhariwal2021diffusion, ramesh2021zero, saharia2022photorealistic}, MUSE adopts a purely transformer-based generation approach outlined in MaskGIT~\cite{chang2022maskgit} with improved inference speed enabled by parallel decoding. In particular, MUSE uses a pre-trained T5-XXL text encoder~\cite{roberts2019exploring} to first encode the text-prompt into a 4096-dimensional language embedding, while the input images are passed through a semantic tokenizer such as VQGAN~\cite{DBLP:conf/cvpr/EsserRO21} consisting of an encoder and a quantization layer to map the image into a sequence of discrete tokens from a learned codebook, along with a decoder which maps the predicted or generated tokens back into the images. 
To enable high resolution image synthesis, MUSE adopts the use of two VQGAN models with different downsampling ratio and spatial resolution of the tokens, along with two transformer modules called \textit{base} and \textit{super-resolution} modules that generate low resolution and high resolution latent tokens respectively. During training, these tokens are trained using cross-entropy loss with a reference image, while during inference, the high resolution latent tokens are passed through the VQGAN decoder to generate images conditioned on the input text prompts. We refer the reader to \cite{chang2023muse} for further details about the training and inference of MUSE generative models.

The use of MUSE in our pipeline as opposed to diffusion-based or auto-regressive based generative models is motivated by two favorable properties. Firstly, MUSE allows efficient mask-based editing capabilities based on its tokenized encoding and decoding-based architecture. Furthermore, MUSE is capable of high-resolution image synthesis with fast inference time and flexible latent token sizes, facilitating the generation of thousands of high quality synthetic post-disaster images conditioned on pre images at scale.

\subsection{Generating Synthetic Data}
\label{subsec:generating_data}

\noindent We provide an overview of our proposed generation pipeline in \cref{fig:generation_fig}. 
We first take the pre-disaster image from the target domain $U_t \in \mathbb{R}^{H \times W \times 3}$, where $H,W$ are the height and width of the image, and generate a corresponding binary mask $\mathcal{M} \in \mathbb{R}^{H \times W}$ with a center patch of size $H',W'$ as ones and rest as zeros, with $H' < H$ and $ W'<W$. Since the aerial images in the datasets we considered are typically centered around the subject, we align the center of the binary mask with the center of the image, helping us in directly editing the relevant subject in the image.
In practice, we randomly perturb the mask around the center by a factor $(\delta_x,\delta_y)$ in both dimensions, where the perturbation factor $\delta_x \sim \mathbb{U}[-W/16,W/16]$ and $\delta_y \sim \mathbb{U}[-H/16,H/16]$ is sampled from the uniform distribution. So our patch with all ones in the binary mask is centered around $(H/2+\delta_y,W/2+\delta_x)$ with respect to the input image $U_t$ and mask $\mathcal{M}$. 

We then pass the target image $U_t$ through the VQGAN encoder to compute the latent tokens for the image, and downsample the binary mask $\mathcal{M}$ to match the spatial resolution of the latent tokens (which is $1/16^{\text{th}}$ and $1/8^{\text{th}}$ of the original image size for the low-resolution and high-resolution pipelines respectively). Subsequently, we multiply the downsampled binary mask $\mathcal{M}'$ with the latent tokens at each pixel location to get a masked token embedding $u_t$.  
The masked-latent token embeddings along with the text-embedding of the input prompt are then passed through a series of cross-attention layers pre-trained in MUSE to predict the output tokens from the predicted latent tokens. The outputs are then passed through the decoder layer of the VQGAN resulting in the output image $\hat{V}_t$. 

We generate four output images for each pre-image and prompt pair, and pick the best one using the ranking obtained by CLIP similarity score~\cite{hessel2021clipscore} between the input prompt and the generated image. We repeat this process for every image in the target dataset, creating a synthetic dataset $\hat{\mathcal{D}}_t = \{U_t^i, \hat{V}_t^i, \hat{y}^i\}_{i=1}^{N_t}$ of pre and post disaster images from the target domain. 



\paragraph{Prompt Pool for Generation}
\noindent A major advantage of generating synthetic images is avoiding the need for manual annotation for unlabeled domains, as the labels can be directly derived from the corresponding prompts.
In our setting with binary labels indicating damaged or not damaged buildings or locations, we choose the prompts to reflect these criterion. For example, to create synthetic images for scenes damaged by \textit{hurricane} disaster, we use prompts such as \textit{An aerial view of a house damaged due to a hurricane} or \textit{A satellite image of a building that was destroyed by a hurricane} and assign the label 1. Alternatively, for generating images which have no damage, we create a prompt pool indicating images which are undamaged (for example, \textit{A satellite image of a building}) and assign the generated images with a label of 0. We list the pool of prompts adopted in our work in \cref{fig:prompt_pool_skai} and \cref{fig:prompt_pool_xbd}. 

\begin{tcolorbox}[colback=red!3,colframe=red!75!white,title=\textsc{SKAI Dataset},left=0.5ex,right=0.5ex,top=0.5ex,bottom=0.5ex]
\fontsize{9}{9}\selectfont
{\bf \underline{Damaged Set}} \\
\emph{An aerial view of a house damaged due to a hurricane.}\\ 
\emph{A bird's-eye view of a building destroyed by a hurricane.}\\
\emph{A top-down view of a house damaged by a hurricane.}\\
\emph{A satellite image of a building destroyed by a hurricane.}\\
\emph{A bird's-eye view of a building damaged by a hurricane.} \\
{\bf \underline{Undamaged Set}} \\
\emph{A satellite image of a house covered by trees.}\\ 
\emph{A bird's-eye view of a house surrounded by trees.}\\
\emph{A top-down view of a house under tree shade.}\\
\emph{An aerial view of an intact house under tree shade.}\\
\end{tcolorbox}
\noindent \begin{minipage}{0.45\textwidth}
\captionof{figure}{Prompt Pool for SKAI}\label{fig:prompt_pool_skai}
\end{minipage}

\begin{tcolorbox}[colback=blue!3,colframe=blue!75!white,title=\textsc{xBD Dataset},left=0.5ex,right=0.5ex,top=0.5ex,bottom=0.5ex]
\fontsize{9}{9}\selectfont
{\bf \underline{Moore Tornado}} \\
\emph{An aerial view of a house damaged due to a tornado.}\\ 
\emph{A bird's-eye view of a building destroyed by a tornado.}\\
\emph{A top-down view of a house damaged by a tornado.}\\
\emph{A satellite image of a building destroyed by a tornado.}\\
\emph{A bird's-eye view of a building damaged by a tornado.}\\
{\bf \underline{Nepal Floods}} \\
\emph{An aerial view of houses surrounded by a flood.}\\ 
\emph{A top-down view of houses damaged by floods.}\\
\emph{A top-down view of a house damaged by floods inundated in water.}\\
\emph{A satellite image of a building destroyed by a flood surrounded by water.}\\
\emph{A satellite image of houses that was destroyed by a flood surrounded by water and trees.}\\
{\bf \underline{Portugal Wildfire}} \\
\emph{An aerial view of forest land after it is torched by a wildfire.}\\ 
\emph{An aerial view of buildings after a wildfire.}\\
\emph{An aerial image of forest land scorched by a wildfire.}\\
\emph{A bird's-eye view of a forest region with completely scorched trees.}
\end{tcolorbox}
\noindent \begin{minipage}{0.45\textwidth}
\captionof{figure}{Prompt Pool for xBD}\label{fig:prompt_pool_xbd}
\end{minipage}

\subsection{Training using Synthetic Data}
\label{subsec:training}

Following prior work in disaster assessment task~\cite{bandara2022transformerbased}, we adopt a siamese network with shared transformer backbone~\cite{dosovitskiy2020image} for training. Specifically, we pass both pre and post images $U$ and $V$ using parameter-shared transformer backbones $\mathcal{E}_u$ and $\mathcal{E}_v$, resulting in feature embedding $f_u=\mathcal{E}_u(U)$ and $f_v=\mathcal{E}_v(V)$ respectively, each dimension $d$. We then fuse these embeddings by concatenating them to form $f \in \mathbb{R}^{2d}$, where $f = \text{concat}(f_s, f_d)$. Finally, we add a 2-layer MLP network $\mathcal{H}$ with a hidden dimension of $d$ to predict a single output value indicating the probability of damaged building between the pre and post images. The whole network is then trained with a binary cross entropy loss using the binary ground truth labels. 

However, directly training predictive models using only synthetic data might result in poor accuracy due to the domain gap between synthetic and real images~\cite{sariyildiz2023fake}. Therefore, we devise a two-stage training strategy to leverage the in-domain synthetic data, along with out-of-domain real data to effectively improve the target performance. In particular, we first train our network end-to-end including the encoders $\mathcal{E}_u$ and $\mathcal{E}_v$ as well as the MLP layers $\mathcal{H}$ using the source domain data $\mathcal{D}_s$. 
%
Subsequently, we follow prior work in robust learning~\cite{kirichenko2022last} to fine-tune only the final layers of the MLP network $\mathcal{H}(.)$ using the synthetic data supervision from the target, while keeping the encoders fixed during the fine-tuning stage. We observed that only re-training the last layer prevents over-fitting the network to the synthetic data compared to complete end-to-end fine-tuning (\cref{tab:model_ablation}), so we adopt this two-stage mechanism in our framework. 
%
During inference, we apply a sigmoid layer on top of the predicted output and threshold this probability to predict damaged buildings in the post-image if the predicted probability is $>0.5$, and predict no damage otherwise.

\paragraph{Fine-tuning the MUSE model} In the framework illustrated so far, we only use a frozen pre-trained MUSE model, where we fix the generative model itself and only use it for inference given input images and corresponding text prompts. However, such off-the-shelf models trained on billions of web-scale image-text data might contain images from a wide variety of domains, and might not be fully suited for use in specific domains like aerial or satellite imagery. Therefore, we also investigate the potential benefits offered by fine-tuning the pre-trained generative model for the specific task of aerial image classification. 
In particular, we collect the pre-disaster images from $\mathcal{D}_t$ and create prompts for each image from the \textit{undamaged} pool to create a dataset of image-text pairs from the target domain. We then adopt adapter-fine tuning~\cite{houlsby2019parameter} to fine-tune the pre-trained model using these image-text pairs, which we found to be more resource-efficient than end-to-end fine-tuning. This fine-tuned model is expected to capture more domain specific properties unique to aerial and satellite imagery, and we compare this procedure with generation using the frozen model in \cref{subsec:results}. 

\section{Experiments}
\label{sec:experiments}

\begin{table*}[htbp]
    \centering
    \resizebox{0.96\textwidth}{!}{
    \begin{tabular}{lcccccccccccccccccc}
        \toprule
        Method && \multicolumn{3}{c}{Ian $\rightarrow$} && \multicolumn{3}{c}{Michael$\rightarrow$} && \multicolumn{3}{c}{Laura$\rightarrow$} && \multicolumn{3}{c}{Maria$\rightarrow$} && Avg. \\
               && Michael & Laura & Maria && Ian & Laura & Maria && Ian  & Michael & Maria && Ian & Michael & Laura && \\
        \midrule
        Source Only && 41.6 & 19.3 & 27.3 && 38.0 & 32.0 & 29.7 && 38.3 & 46.9 & 26.3 && 30.0 & 39.6 & 21.9 && 32.6 \\
        Ours w/ ZeroShot MUSE && 49.2 & \textbf{36.8} & \textbf{31.9} && 47.4 & \textbf{42.5} & \textbf{32.0} && \textbf{50.0} & 54.7 & \textbf{30.6} && 42.6 & \textbf{54.5} & \textbf{36.6} && \textbf{42.4} (\textcolor{violet}{+9.8\%})  \\
        Ours w/ fine-tuned MUSE && \textbf{49.6} & 29.9 & 25.8 && \textbf{50.9} & 31.5 & 28.8 && 49.2 & \textbf{55.6} & 26.4 && \textbf{44.1} & 53.6 & 27.6 && 39.4 (\textcolor{violet}{+6.8\%}) \\
        \bottomrule
    \end{tabular}
    }
    \captionsetup{width=\textwidth}
    \caption{{\bf Single-source Domain Adaptation Results on SKAI dataset} AUPRC values for different transfer settings from the SKAI dataset. We compare the results obtained by training using only real data from the source domain and combining it with synthetic generated data from the target domains on all the transfer settings. Evidently, our approach outperforms the source-only baseline setting new state-of-the-art.}
    \label{tab:skai_uda_table}
\end{table*}

\begin{table*}[htbp]
    \centering
    \resizebox{0.9\textwidth}{!}{
    \begin{tabular}{lcccccccccccccccccc}
        \toprule
        Method && \multicolumn{2}{c}{Moore-Tornado$\rightarrow$} && \multicolumn{2}{c}{Nepal-Flooding$\rightarrow$} && \multicolumn{2}{c}{Portugal-Wildfire$\rightarrow$} && Avg. \\
               && Nepal-Flooding & Portugal-Wildfire && Moore-Tornado & Portugal-Wildfire && Moore-Tornado & Nepal-Flooding && \\
        \midrule
        Source Only && 23.8 & 23.2 && 14.5 & 18.5 && 45.3 & 24.7 && 25.0 \\
        Ours w/ ZeroShot MUSE && \textbf{49.5} & \textbf{24.1} && 75.1 & 25.1 && 76.0 & \textbf{51.6} && 50.2 (\textcolor{violet}{+25.2\%})  \\
        Ours w/ fine-tuned MUSE && 43.9 & \textbf{24.1} && \textbf{82.3} & \textbf{25.3} && \textbf{83.1} & 47.5 && \textbf{51.1} (\textcolor{violet}{+26.1\%}) \\
        \bottomrule
    \end{tabular}
    }
    \captionsetup{width=\textwidth}
    \caption{{\bf Single-source Domain Adaptation Results on xBD dataset.} AUPRC values for different transfer settings from the xBD dataset~\cite{gupta2019xbd}. On each of the transfer setting, augmenting training using synthetic data from the target domain significantly outperforms the source-only baseline, with an improvement of 25.2\% using a zeroshot generative model, and 26.1\% with further fine-tuning the generative backbone on aerial image-text pairs.}
    \label{tab:xbd_uda_table}
\end{table*}

We next demonstrate the effectiveness of the proposed approach on several challenging transfer settings. We first introduce our choice of datasets in \cref{subsec:datasets}, specify the training details in \cref{subsec:training_details} followed by the results in \cref{subsec:results} and several ablations into our modeling and training choices in \cref{subsec:ablations}. 

\begin{table*}[htbp]
    \centering
    \resizebox{\textwidth}{!}{
    \begin{tabular}{lcccccccccccccccccc}
        \toprule
        Dataset && \multicolumn{5}{c}{SKAI-Dataset} && \multicolumn{4}{c}{xBD-Dataset} \\
        \cline{3-7} \cline{9-12}
        Method && $\rightarrow$Ian & $\rightarrow$Michael & $\rightarrow$Maria & $\rightarrow$Laura & Avg. && $\rightarrow$Moore-Tornado & $\rightarrow$Nepal-Flooding & $\rightarrow$Portugal-Wildfire & Avg. \\
        \midrule
        Source Only && 44.21 & 48.62 & 29.54 & 36.83 & 39.78 && 18.96 & 27.16 & 29.69 & 25.27 \\
        Ours w/ ZeroShot MUSE && \textbf{54.79} & 52.24 & \textbf{34.05} & \textbf{39.38} & \textbf{45.11} (\textcolor{violet}{+5.33\%}) &&  78.70 & \textbf{52.40} & \textbf{32.10} & \textbf{54.40}(\textcolor{violet}{+29.13\%}) \\
        Ours w/ fine-tuned MUSE && 49.12 & \textbf{53.83} & 30.92 & 39.29 & 43.29 (\textcolor{violet}{+3.51\%}) && \textbf{83.18} & 50.16 & 30.72 & \textbf{54.69}(\textcolor{violet}{+29.42\%}) \\
        \bottomrule
    \end{tabular}
    }
    \captionsetup{width=\textwidth}
    \caption{{\bf Multi-source Domain Adaptation Results on SKAI and xBD datasets.} AUPRC values for different transfer settings, where we show the result of training using synthetic generated data from the respective target domain along with manual supervision from all the three remaining domains. Our approach outperforms the source-only baseline highlighting the effectiveness of training with generated synthetic data in bridging domain gaps.}
    \label{tab:loo_table}
\end{table*}

\subsection{Datasets}
\label{subsec:datasets}

\paragraph{xBD Dataset}
xBD \cite{gupta2019xbd} is a large-scale dataset designed for automatic disaster assessment using aerial and satellite imagery. The dataset covers synchronized pre- and post- event satellite imagery of both damaged and undamaged scenes from more than 19 events across the world, covering a variety of disaster types across varying severity levels. Since our focus in this paper is to improve robustness of aerial disaster assessment algorithms across disparate geographies, we choose 3 domains from xBD, namely \textit{nepal-flooding}, \textit{portugal-wildfires} and \textit{moore-tornado} to demostrate our results, which have 36456, 18884 and 18491 images respectively. These domains encompass data from three distinct geographical subregions, each affected by entirely different types of disasters making it a challenging problem to improve cross-domain robustness. 


\paragraph{SKAI Satellite Imagery} In order to verify the effectiveness of our method in improving the performance across subtle domain variations, we adopt the SKAI dataset~\cite{lee2020assessing} consisting of pre and post hurricane images captured from different regions in the United States. The images in SKAI includes data collected from Ian, Maria, Michael and Laura hurricanes with 2733, 3709, 3991 and 3991 images respectively, which we use as the different domains for our cross-domain robustness setting. Note that both these datasets consist of heavy class imbalance, with more than 80\% of the image-pairs capturing non-damaged buildings, adding an additional layer of complexity in bridging the domain shifts. For both the datasets, we show results using single-source and multi-source adaptation settings, in which we use supervised data from single source domain or all the domains except the target respectively.

\subsection{Training and Evaluation Details}
\label{subsec:training_details}

\noindent We use an Imagenet pretrained ViT-B/16 transformer backbone~\cite{steiner2021train} as the encoder in our setting, and remove the last classification layer replacing it with the MLP head for binary classification. We then train the network using the two-stage approach discussed in \cref{subsec:training}, first using the supervised source domain images using Adam optimizer with a learning rate of 2e-6 for the pre-trained backbone and 2e-5 for the randomly initialized MLP layer, followed by re-training only the last MLP layer using synthesized target domain images using the same hyperparameters as above. We use a batch size of 64 in both cases and perform training for 5000 iterations. We use the validation images from the target domain to perform early stopping, which we observed to be very crucial to obtain good performance in our setting. 

Following prior work in disaster assessment tasks from satellite imagery~\cite{gupta2019xbd, lee2020assessing}, we adopt the AUPRC metric for evaluation which measures the area under the precision-recall curve across various thresholds, and is shown to be relatively more robust for cases like ours where there is severe class imbalance against positive examples. In terms of baselines, we compare with a source-only baseline which only trains a predictive model on the source domain and evaluates on the target test-set. Since this does not use any target data, it serves as a fundamental baseline to illustrate the benefits obtained by our method. Note that prior UDA methods require both pre and post disaster images to learn domain agnostic features~\cite{benson2020assessing, Li2020UnsupervisedDA, Valentijn2020transferability}, preventing a direct comparison for our setting where only pre-disaster images from the target dataset are available. 

\subsection{Results}
\label{subsec:results}

\paragraph{Single-source Zeroshot Adaptation} We show the results for single-source UDA for domains from the SKAI dataset in \cref{tab:skai_uda_table} and xBD dataset in \cref{tab:xbd_uda_table}. As shown, our method of augmenting out of distribution training using synthetic images generated from MUSE model achieves better accuracy than the source-only baseline, with $\sim 10\%$ and $\sim 25\%$ improvements on the SKAI and the xBD datasets on average. Our improvements are consistent across all the transfer settings, with up to $\sim 70\%$ improvement on the more challenging cross-disaster cross-geography setting from xBD dataset, highlighting the effectiveness of leveraging generative foundational models to create synthetic data for low-resource domains even in expert tasks like disaster assessment.

Furthermore, we also compare the AUPRC results observed through fine-tuning the generative model on aerial imagery and satellite images, using the procedure outlined in \cref{subsec:training}. We observe that the model trained with data generated from fine-tuned model outperforms both the source-only baseline as well as the zeroshot settings on 4 out of 6 settings in xBD dataset with $\sim1\%$ improvement on the average accuracy, indicating the potential in fine-tuning MUSE model on domain-specific images. On SKAI data however, we observe zeroshot model is better on the average AUPRC.
A potential reason for this could be that the generative ability of the text-to-image generative model is reduced after fine-tuning on domain-specific images, impacting accuracy in few of the transfer settings, highlighting room for further improvement through more carefully designed fine-tuning strategies. 

\paragraph{Multi-source Zeroshot Adaptation} The comparison for both SKAI and xBD datasets on multi-source adaptation setting is shown in \cref{tab:loo_table}. Firstly, the accuracy achieved by multi-source models on all target domains is higher than single-source setting, which is expected since multi-source models have access to relatively more supervised data. Furthermore, the results from \cref{tab:loo_table} clearly show the effectiveness of our approach even for such multi-source evaluation setting, where our method using zeroshot text-to-image generation yields $5.33\%$ improvement over baseline on SKAI dataset and $29.13\%$ improvement over baseline on the xBD dataset. Our benefits are consistent for both the datasets across all the transfer tasks, further supporting our hypothesis that text-to-image models can serve as strong data generators for low-resource domains. As seen for the case of single-source setting, we observe the gains yielded by data generation using zeroshot text-to-image models to be competitive when compared to fine-tuned models on both the datasets.

\subsection{Ablations}
\label{subsec:ablations}

\begin{table}[!t]
    \centering
    \resizebox{0.48\textwidth}{!}{
    \begin{tabular}{lcccccccc}
        \toprule
        Method && SKAI && xBD \\
        \midrule
        (\texttt{R0}) Source Only && 39.78 && 25.27 \\
        (\texttt{R1}) Only Synthetic Data && 40.60 && 47.76  \\
        (\texttt{R2}) Joint Training on Real + Synthetic && 43.11  && 49.66\\
        (\texttt{R3}) End-to-end Finetuning && 44.10  && 53.44  \\
        \midrule
        (\texttt{R4}) Last-Layer Finetune on SynData && \textbf{45.11} && \textbf{54.40} \\
        \bottomrule
    \end{tabular}
    }
    \captionsetup{width=0.48\textwidth}
    \caption{{\bf Effect of training choices} We show the effect of various training choices in our framework, where last-layer re-training using only synthetic data (\texttt{R4}) outperforms training using only synthetic data without the source labels (\texttt{R1}), jointly training on both real and synthetic data (\texttt{R2}) as well as end-to-end finetuning using synthetic data (\texttt{R3}).}
    \label{tab:model_ablation}
\end{table}

\paragraph{Ablations and Insights} We show the effect of various design choices in our framework in \cref{tab:model_ablation}. Firstly, we observe that training only using synthetic data without source domain data leads to poor results, potentially highlighting the limitations of synthetic data alone in training (\texttt{R1} vs \texttt{R4}). This facet of synthetic data has also been noted in prior works~\cite{sariyildiz2023fake}, indicating that manual supervision is still necessary to observe gains with synthetic supervision. Furthermore, we also show that joint-training using real source and synthetic target datasets is inferior to our proposed approach of first pre-training on real source data followed by last-layer retraining on generated target domain data (\texttt{R2} vs \texttt{R4}), supporting our two-stage training framework. Finally, we also observe that end-to-end fine-tuning using synthetic data under-performs the approach of finetuning the last layer only (\texttt{R3} vs \texttt{R4}).

\begin{figure}[!tpb]
\centering
\begin{minipage}[t]{0.46\linewidth}
    \centering
    \includegraphics[width=\textwidth]{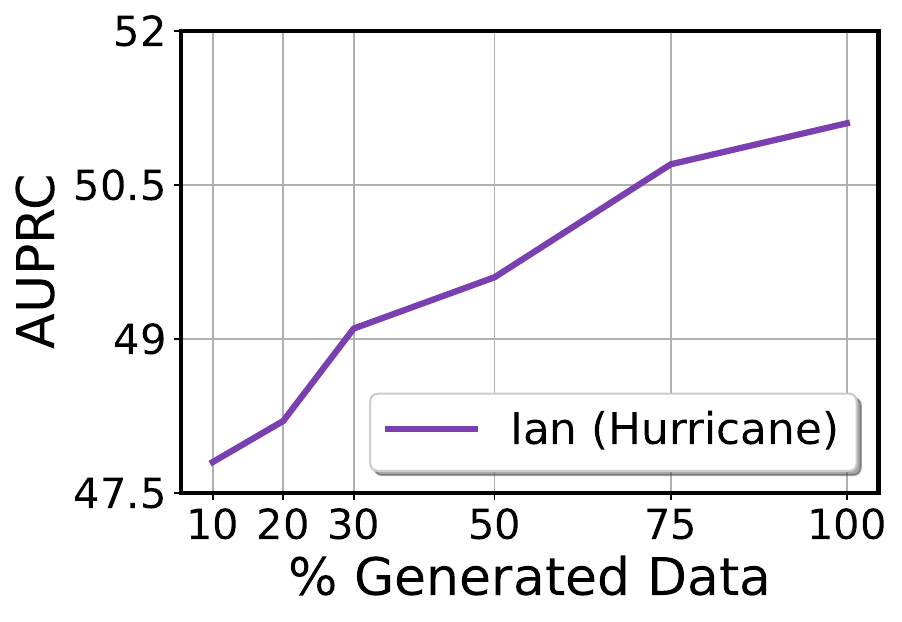}
    \subcaption{Ian Hurricane}
    \label{fig:ian_hurricane}
\end{minipage}
~
\begin{minipage}[t]{0.46\linewidth}
    \centering
    \includegraphics[width=\textwidth]{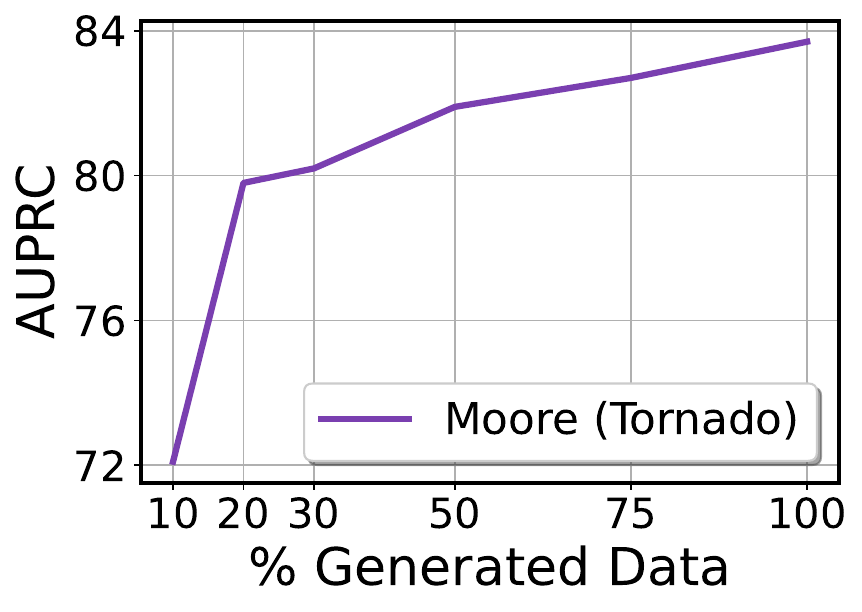}
    \subcaption{Moore Tornado}
    \label{fig:xbd_moore}
\end{minipage}
\caption{{\bf Effect of the amount of generated synthetic data.} We show the positive influence of the volume of generated synthetic data (as a \% of the target domain images) on the multi-source transfer setting, where adding more target data invariably helps to improve the final target accuracy for both SKAI \subref{fig:ian_hurricane} and xBD \subref{fig:xbd_moore} datasets, with potential for further enhancement with more generated data.}
\label{fig:datavol}
\end{figure}

\paragraph{Effect of Volume of Synthetic Data } We show the effect of the amount of generated synthetic data on the target AUPRC in \cref{fig:datavol}. We observe that adding more synthetic data invariably helps the final target accuracy for both the datasets studied. More importantly, we observe no saturation even when using all target data to generate images indicating further room for improvement of target performance through low-cost synthetically generated data. 

\begin{figure*}[!thpb]
\centering

\begin{minipage}[t]{\textwidth}
    \centering
    \includegraphics[width=0.9\textwidth]{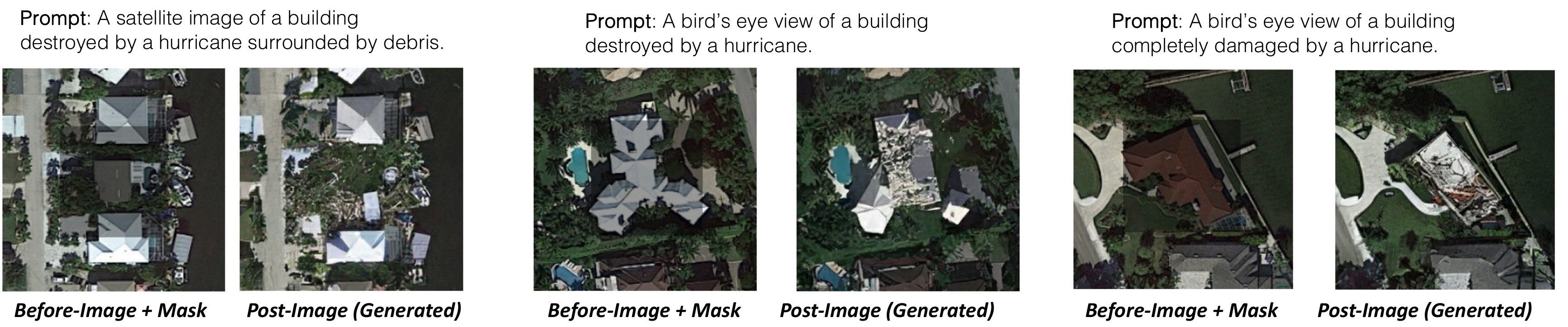}
    \subcaption{Images from Ian domain.}
    \label{fig:skai_vis_ian}
\end{minipage}

\begin{minipage}[t]{\textwidth}
    \centering
    \includegraphics[width=0.9\textwidth]{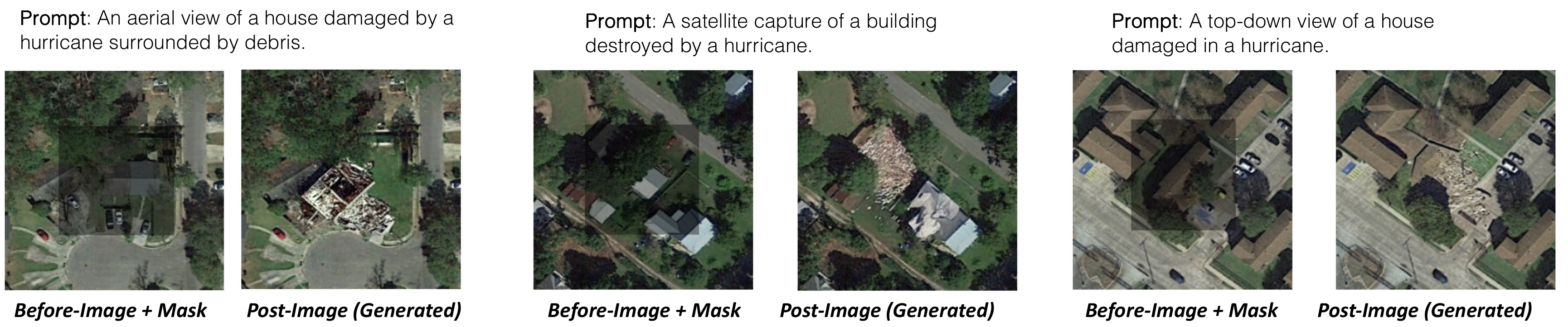}
    \subcaption{Images from the Michael domain.}
    \label{fig:skai_vis_michael}
\end{minipage}

\begin{minipage}[t]{\textwidth}
    \centering
    \includegraphics[width=0.9\textwidth]{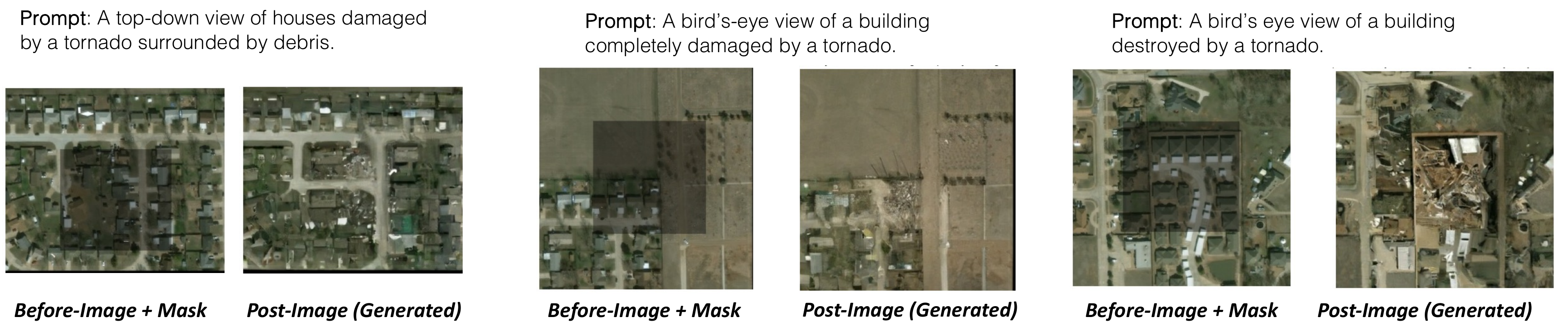}
    \subcaption{Images from Moore-Tornado.}
    \label{fig:xbd_vis_tornado}
\end{minipage}

\begin{minipage}[t]{\textwidth}
    \centering
    \includegraphics[width=0.9\textwidth]{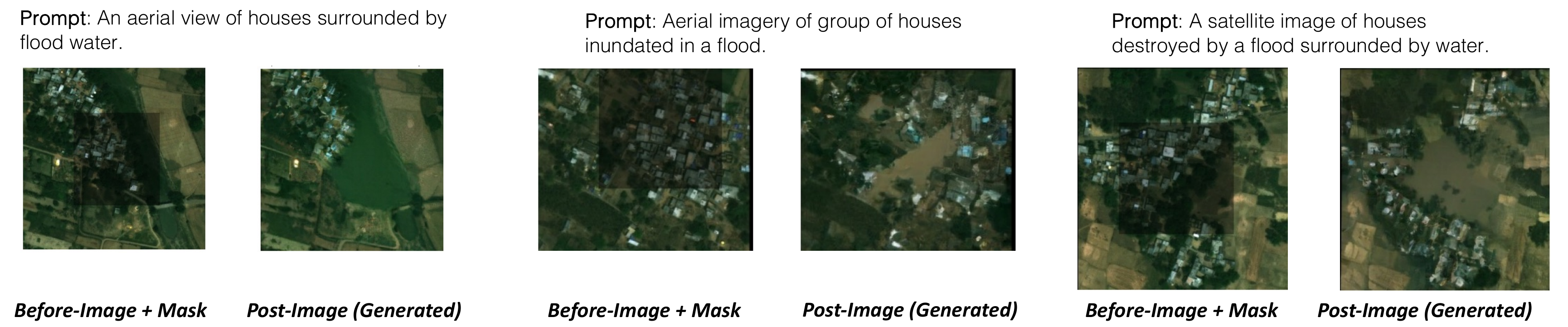}
    \subcaption{Images from Nepal-Flooding.}
    \label{fig:xbd_vis_floods}
\end{minipage}

\begin{minipage}[t]{\textwidth}
    \centering
    \includegraphics[width=0.9\textwidth]{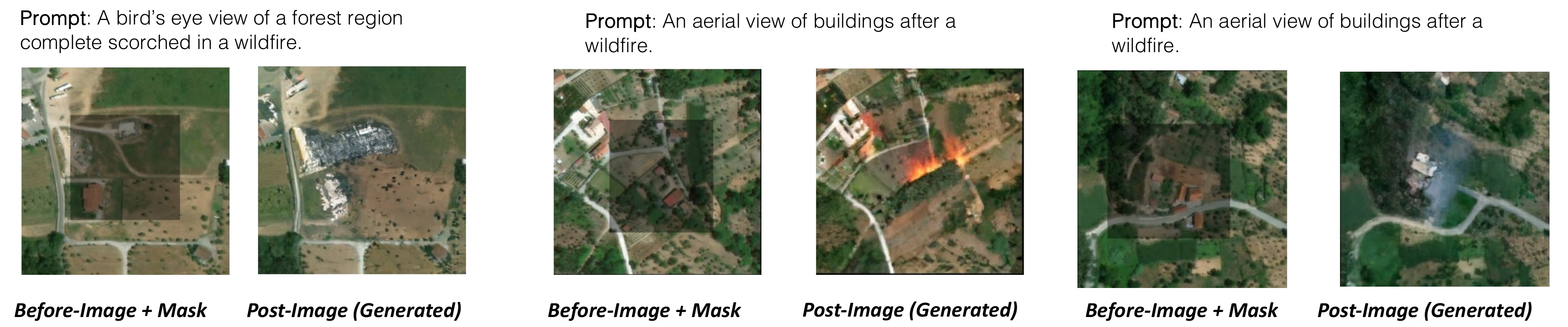}
    \subcaption{Images from Portugal-Wildfire.}
    \label{fig:xbd_vis_wildfire}
\end{minipage}

\caption{\textbf{Visualization of text-to-image results.} We show several examples from our generated images, along with the pre-disaster image, correponding conditioning mask (overlapped on the pre-image) as well as the text-prompt used to generate the post-image from \subref{fig:skai_vis_ian} Ian-Hurricane, \subref{fig:skai_vis_michael} Michael-Hurricane, \subref{fig:xbd_vis_tornado} Moore-Tornado, \subref{fig:xbd_vis_floods} Nepal-Floods and \subref{fig:xbd_vis_wildfire} Portugal-Wildfire. }
\label{fig:visualization}
\end{figure*}

\paragraph{Visualizing Generated Images} We show several illustrations of samples generated through our method in \cref{fig:visualization} on both SKAI (\cref{fig:skai_vis_ian}, \cref{fig:skai_vis_michael}) and xBD (\cref{fig:xbd_vis_tornado}, \cref{fig:xbd_vis_floods}, \cref{fig:xbd_vis_wildfire}) datasets, where we include the pre-disaster image as well as the mask and the text-prompt used for conditional image editing through our generative model. In most cases, we observe that the text-to-image model incorporates the textual guidance and performs localized editing on the input image to generate a synthetic post-disaster image with great effectiveness. The model shows excellent capability in seamlessly handling the various types of disasters through our text-guidance, which helps to create realistic images in low-resource domains leading to significant empirical gains (\cref{tab:loo_table}). 

\section{Conclusion}

In this paper, we explore the potential of leveraging emerging text-to-image models in generating synthetic supervision to improve robustness across low-resource domains for disaster assessment tasks. We design an efficient and scalable data-generation pipeline by leveraging the localized image editing capabilities of transformer-based generative models~\cite{chang2023muse}. Using this framework, we generate several thousand synthetic post-disaster images conditioned on pre-disaster images and text guidance, followed by a simple two-stage training mechanism that yields non-trivial benefits over a source-only baseline in both single source and multi-source domain adaptation setting. 
In terms of limitations, we noted a significant sensitivity of the training process to the quality and coherence of the generated synthetic data, which is directly affected by the presence of low-quality generated images. A potential future work can therefore be to additionally incorporate better filtering strategies into our framework to remove poor quality images and improve training. 
Nevertheless, our work serves as one of the first to explore the potential of text-to-image synthetic data for expert tasks like satellite disaster assessment, which holds massive potential for continued improvement with the development of better image generation models.

{
    \small
    \bibliographystyle{ieeenat_fullname}
    \bibliography{main}
}


\end{document}